\documentclass{article}

 \usepackage[dblblindworkshop,final]{neurips_2025}

\usepackage[utf8]{inputenc} 
\usepackage[T1]{fontenc}    
\usepackage{hyperref}       
\usepackage{url}            
\usepackage{booktabs}       
\usepackage{amsfonts}       
\usepackage{nicefrac}       
\usepackage{microtype}      
\usepackage{xcolor}         
\usepackage{enumitem}
\usepackage{graphicx}

\title{Bridging Tool Dependencies and Domain Knowledge: A Graph-Based Framework for In-Context Planning}

\author{%
  Shengjie Liu \\
  Amazon\\
  \texttt{zycjlsj@amazon.com} \\
  \And
  Li Dong \\
  Amazon \\
  \texttt{ldonga@amazon.com} \\
  \AND
  Zhenyu Zhang \\
  Amazon \\
  \texttt{zhenyuzh@amazon.com} \\
}

\begin{document}

\maketitle

\begin{abstract}
  We present a framework for uncovering and exploiting dependencies among tools and documents to enhance exemplar artifact generation. Our method begins by constructing a tool knowledge graph from tool schemas—including descriptions, arguments, and output payloads—using a DeepResearch-inspired analysis. In parallel, we derive a complementary knowledge graph from internal documents and SOPs, which is then fused with the tool graph. To generate exemplar plans, we adopt a deep–sparse integration strategy that aligns structural tool dependencies with procedural knowledge. Experiments demonstrate that this unified framework effectively models tool interactions and improves plan generation, underscoring the benefits of linking tool graphs with domain knowledge graphs for tool-augmented reasoning and planning.
\end{abstract}

\section{Introduction}



Large Language Models (LLMs) (\cite{brown2020languagemodelsfewshotlearners, JMLR:v24:22-1144,deepseekai2025deepseekr1incentivizingreasoningcapability,touvron2023llama2openfoundation,zeng2023glm130bopenbilingualpretrained}) have played a central role in driving progress in artificial intelligence, yielding major advances across a wide range of domains. A key strength of LLMs lies in their planning abilities and their capacity for tool use (\cite{yao2023treethoughtsdeliberateproblem, yao2023reactsynergizingreasoningacting}), which allow them not only to execute instructions and carry out web queries but also to significantly enhance their mathematical reasoning skills. LLM Compiler \cite{kim2024llmcompilerparallelfunction} and subsequent studies (\cite{erdogan2024tinyagentfunctioncallingedge, erdogan2025planandactimprovingplanningagents}) propose modeling tool interactions as directed acyclic graphs (DAGs), enabling the parallel execution of independent tools and thereby improving efficiency in tool calling. CodeAct \cite{wang2024executablecodeactionselicit} and CodePlan \cite{wen2024unlockingreasoningpotentiallarge} advocate for generating pseudo-Python code as a means of structuring high-level reasoning steps, where each tool invocation is expressed as a function call. ReWOO \cite{xu2023rewoodecouplingreasoningobservations} introduces a modular approach that separates the reasoning workflow from external observations of tool outputs, which reduces token usage and improves efficiency. \cite{liu2025toolplannertaskplanningclusters} groups tools into clusters, plans at the toolkit level, and performs replanning within the same toolkit when errors occur. \cite{ma2024nonmyopicgenerationlanguagemodels} presents Predictive-Decoding, a technique inspired by Model Predictive Control, designed to mitigate early-stage errors and encourage non-myopic planning, ultimately improving accuracy. ReasonFlux \cite{yang2025reasonfluxhierarchicalllmreasoning} outlines a framework where the LLM reasons over template slots, executes tools through these templates, and employs reinforcement learning with action completion rewards to refine planning accuracy.

In the agentic paradigm, LLMs are transitioning from purely text-based reasoning toward dynamic agents capable of planning, tool usage, and multi-step (including multi-turn) execution. ToolRL \cite{qian2025toolrlrewardtoollearning} provides a systematic study of reinforcement learning reward design—examining granularity, temporal structure, and signal types—to enhance generalization in multi-turn, tool-integrated reasoning. Kimi K2 \cite{kimiteam2025kimik2openagentic} demonstrates that stabilizing long-context training combined with multi-stage RL enables strong performance across multi-round tasks in software engineering, mathematics, and agentic reasoning.

Prior work on tool-using capabilities has largely centered on general-purpose tools \cite{qin2023toolllmfacilitatinglargelanguage}, such as \textit{send email} and \textit{make calendar}, as well as web search functionalities exemplified by systems like Manus. These tools are typically uni-functional, designed to perform a single action or answer a single query, with descriptions and argument structures that are straightforward for LLMs such as GPT-4o \cite{openai2024gpt4ocard} and Claude 3.5 to interpret. In contrast, the tools required by real-world business assistants—covering areas such as inventory tracking, performance monitoring, and financial reporting—are far more complex and domain-specific, posing significant challenges for general-purpose LLMs to parse and utilize effectively. To address this gap, in-context planning (ICP) has emerged as a common strategy, where exemplar API executions are provided to LLMs to guide plan generation.

Generating high-quality in-context planning exemplars is essential, as irrelevant or poorly chosen examples can misguide the LLM and lead to suboptimal or erroneous plans \cite{zhao2025improvinglargelanguagemodel}. For a business assistant that must initially operate over hundreds of in-domain tools, obtaining sufficient and high-quality exemplars is crucial for overcoming the cold-start problem in in-context planning and ensuring reliable functionality. Therefore, it is highly important to automatically construct a tool knowledge graph that captures the dependencies among tools and links them to internal documents or SOPs, which provide the corresponding usage instructions.

In summary, this work makes several pivotal contributions:
\begin{itemize}[left=0pt, itemsep=0pt]
    \item We propose a Deep Research approach \cite{xu2025comprehensivesurveydeepresearch} to explore tool schemas—including descriptions, arguments, and output payloads—in order to uncover dependencies among tools and construct a tool knowledge graph. 
    \item We further model internal documents and SOPs containing tool usage instructions as a knowledge graph via GraphRAG \cite{edge2025localglobalgraphrag}, and integrate this with the tool graph through knowledge graph fusion. Based on this fusion graph, we propose a Dense-sparse integration framework following HippoRAG2 for exemplar plan generation. 
    \item Experimental results validate the effectiveness of studying tool dependencies with deep research and demonstrate the usefulness of connecting tool graphs with document knowledge graphs. 
\end{itemize}

\section{METHODOLOGY}
In this section, we describe how to adapt Deep Research—originally developed for web search and report generation—to the task of exploring dependencies among tools using their schemas. We then propose an appropriate data structure to represent and store these dependencies. Next, we explain how GraphRAG can be leveraged to construct a domain knowledge graph from internal documents, and how the two graphs can be integrated through knowledge fusion.

\subsection{Tool Graph Construction}
We modify the node in Deep Research original pipeline (web-search and then refine with LLM feedback) into dependency extraction and then use LLM as a judge \cite{gu2025surveyllmasajudge} to check if the identified the dependency really makes sense in the sense we delete those which should not form the dependency per in-domain specific requirements. When send pairwise tools to LLM and let it check the tools description, input arguments and also the output payload. The extracted dependency data structure is shown in Figure \ref{fig:image1}. The pipeline is written in LangGraph.

\begin{figure}[!ht]
  \centering
  \includegraphics[width= 0.7\linewidth]{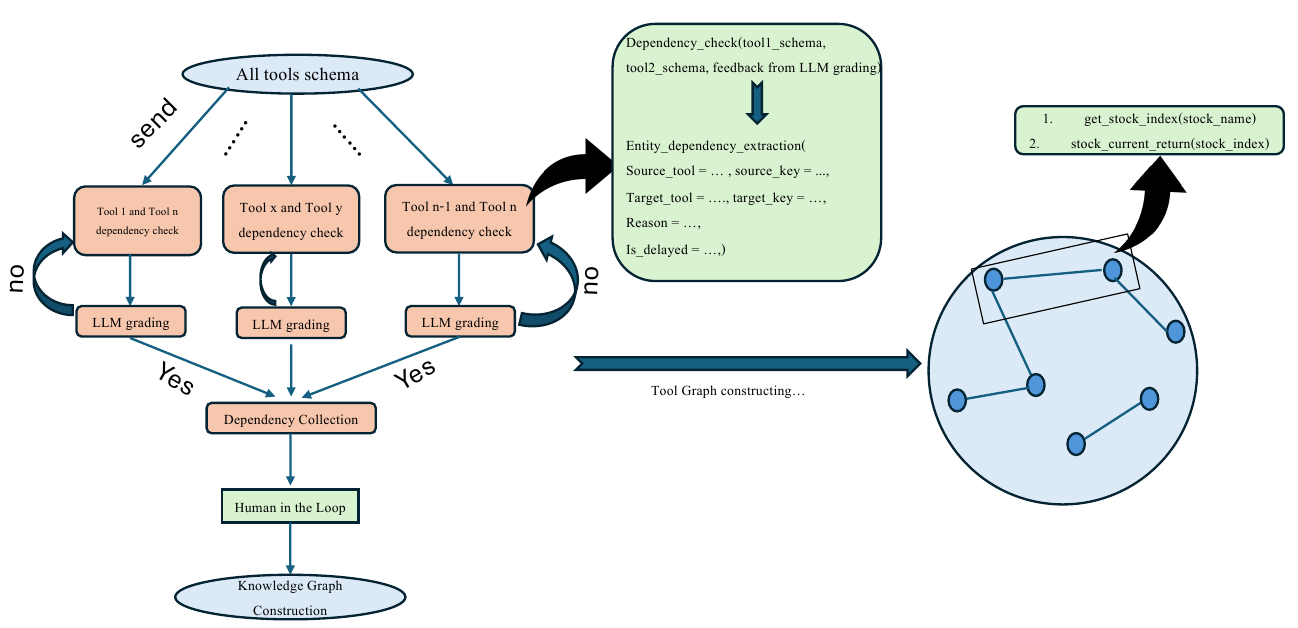} 
  \caption{Tool Graph Construction}
  \label{fig:image1}
\end{figure}

\subsection{Fusion of Tool Graph and Domain Knowledge Graph}
We first construct the domain knowledge graph, aligned with the corresponding internal documents and SOPs, using GraphRAG\footnote{https://github.com/awslabs/graphrag-toolkit} with its default setup. We then enrich this graph by incorporating the tool graph through graph fusion in Neptune\footnote{https://aws.amazon.com/neptune/}, where the dependency relation is defined as \texttt{\_can\_use\_this\_tool\_output}.

After constructing the unified knowledge graph that integrates the tool graph, we develop a pipeline to derive the final exemplar artifacts for In-Context Planning (ICP) from this unified representation. We adopt HippoRAG2 \cite{gutiérrez2025ragmemorynonparametriccontinual} as the pipeline for generating plan artifacts, as the integration of domain knowledge with the tool graph aligns with the Dense–Sparse Integration framework described in HippoRAG2, where the tool graph serves as the sparse component and the domain knowledge represents the dense component.

\begin{figure}[h]
  \centering
  \includegraphics[width= 0.7\linewidth]{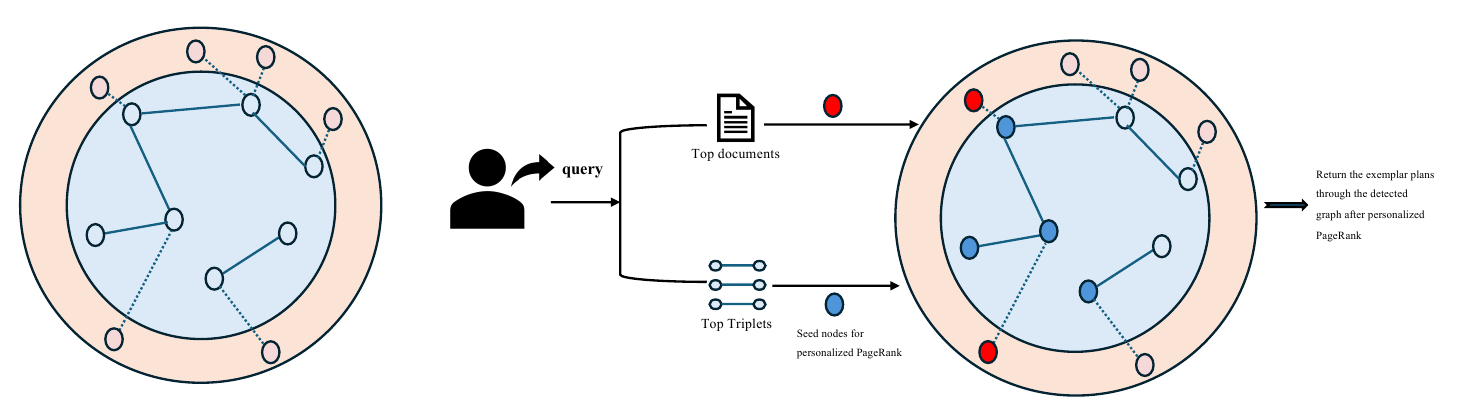} 
  \caption{Dense-Sparse Integration framework for exemplar plans generation. Red dots represent document nodes; blue dots represent tool nodes.}
  \label{fig:image1}
\end{figure}

To prepare exemplar artifacts for cold-start ICP, we begin by collecting a set of queries from production. For each query, we retrieve the top-K tool triplets and relevant knowledge documents using embedding search. The selected tool triplets are then used as seed nodes to run Personalized PageRank, producing a subgraph tailored to the query. Leveraging this subgraph, LLM generates the exemplar artifacts, which are subsequently stored in the vector database for future retrieval.

\section{EXPERIMENTS}
Similar to prior work \cite{liu2025toolplannertaskplanningclusters}, we adopt \textsc{ToolBench} \cite{qin2023toolllmfacilitatinglargelanguage} as our benchmark dataset. \textsc{ToolBench} includes 16,464 APIs and provides three levels of prompts—$G_1$, $G_2$, and $G_3$—for generating queries and corresponding plans using depth-first search (DFS) planning. Specifically, $G_1$ corresponds to single-tool instructions, $G_2$ to intra-category multi-tool instructions, and $G_3$ to intra-collection multi-tool instructions. We repurpose the \textsc{ToolBench} data by randomly selecting 1,000 queries from $G_1$, 1000 from $G_2$, and 1000 from $G_3$—and use the ground-truth provided in the dataset \label{data}. After filtering out invalid cases, we identified 1,500 valid tool dependencies derived from the queries and their corresponding ground-truth plans.
To evaluate whether our Dense–Sparse Integration framework improves exemplar plan generation, we simulate external knowledge by using the Tavily-Search API. For each API output payload in our dataset, we retrieve documents online that can utilize the corresponding output payloads and add the documnents instruction in the ground-truth artifacts.

\subsection{Performance on Dependency Checking}
We apply the dependency-checking pipeline to all API documentation in our dataset and evaluate performance across a range of LLMs. As shown in Table~\ref{tab:freq0}, both GPT-4o and Claude-4 achieve strong results, with precision and recall rates exceeding 80\%. Notably, despite its smaller model size, Qwen3-8B also performs well and even attains slightly higher precision than Claude-4. The performance across most models demonstrates the validity of our framework in accurately identifying dependencies.
\begin{table}[ht]
\centering
  \caption{Performance on Dependency Checking}
  \label{tab:freq0}
  \resizebox{0.5\linewidth}{!}{%
  \begin{tabular}{ccccccc}
    \toprule
    Model & Predicted Dependencies & True Dependencies & Precision & Recall \\
    \midrule
    GPT-4o & 1332& 1500&\textbf{90.7\%} & 80.5\% \\
    Claude 4 & 1652& 1500&79.9\% & \textbf{88.1\%} \\
     Claude 3.7 & 1462&1500& 80.9\% & 78.9\%\\
       DeepSeek R1 &1652&1500& 66.4\% & 73.2\% \\
       Qwen3-8B &1453& 1500&83.2\% & 80.7\% \\
    \bottomrule
  \end{tabular}%
  }
\end{table}
\subsection{Dense-sparse Framework for Exemplar Plan Generation}
In this section, we assess the effectiveness of our Dense–Sparse framework under the setting where the unified knowledge graph correctly integrates tools and domain knowledge. To this end, we begin with the 3,000 queries introduced in Section~\ref{data}. Since ground-truth plans are available for each query, we employ LLM-as-a-judge to verify whether the generated exemplar artifacts align with the ground-truth. We employ Jina Embedding (v3)~\cite{sturua2024jinaembeddingsv3multilingualembeddingstask} to compute semantic similarity, which is then used to retrieve the top-ranked documents and tool triplets, and we use the same set of LLMs as in Table~\ref{tab:freq0} for generating the final plan artifacts after applying Personalized PageRank. The range of LLM-as-a-judge score ranges from 0 to 2 for the plan coverage. We use Nova pro \footnote{https://aws.amazon.com/ai/generative-ai/nova/} as the LLM judge. 
\begin{table}[ht]
\centering
  \caption{Performance on Exemplar Plan Generation}
  \label{tab:freq1}
  \resizebox{0.5\linewidth}{!}{%
  \begin{tabular}{ccccccc}
    \toprule
    Model & Binary Match Accuracy & LLM-as-a-judge Score \\
    \midrule
    GPT-4o & 77\% & 1.62\\
    Claude 4 & 69\% & 1.47 \\
     Claude 3.7 & 71\% & 1.49\\
       DeepSeek R1 & 64\% & 1.36 \\
       Qwen3-8B & 72\% & 1.58 \\
    \bottomrule
  \end{tabular}%
  }
\end{table}
Based on both binary match accuracy and LLM-as-a-judge scores, we observe that performance remains relatively stable across models. Notably, Qwen3-8B achieves strong results despite its smaller size. This consistency suggests that performance may be largely determined by the quality of the subgraph returned by Personalized PageRank and embedding match, rather than solely by model capacity. We also conduct an ablation study to evaluate the effectiveness of Personalized PageRank by comparing performance with and without its application. 
\begin{table}[ht]
\centering
  \caption{Abalation study on Personalized PageRank in the pipeline}
  \label{tab:freq1}
  \resizebox{0.5\linewidth}{!}{%
  \begin{tabular}{ccccccc}
    \toprule
    Model & Binary Match Accuracy & LLM-as-a-judge Score \\
    \midrule
    GPT-4o with Personalized PageRank & 77\% & 1.62\\
    GPT-4o without Personalized PageRank & 68\% & 1.56 \\
    \bottomrule
  \end{tabular}%
  }
\end{table}
Removing Personalized PageRank yields a ~9 percentage-point drop in binary match accuracy. A deeper analysis shows that PPR recovers tool documents and procedural instruction pages that embedding-only retrieval often overlooks, enriching the subgraph used for plan generation. For example, pure embedding search can miss tools like BacklogCheck when they are dominated by other inventory-related tools, whereas PPR propagates importance through dependency links to surface them.


\section{CONCLUSION}
In conclusion, we introduced a framework for uncovering and leveraging dependencies between tools and documents to improve exemplar artifact generation. By constructing a tool knowledge graph from schemas and fusing it with a domain knowledge graph from internal documents and SOPs, our deep–sparse integration strategy aligns structural tool dependencies with procedural knowledge. Experiments confirm that this unified approach effectively models tool interactions and enhances plan generation. Nonetheless, the absence of real benchmarks for detecting tool dependencies remains a limitation, which we plan to address in future work.
\newpage

\bibliography{nips}

\begin{thebibliography}{10}

\bibitem{brown2020languagemodelsfewshotlearners}
Tom~B. Brown, Benjamin Mann, Nick Ryder, Melanie Subbiah, Jared Kaplan, Prafulla Dhariwal, Arvind Neelakantan, Pranav Shyam, Girish Sastry, Amanda Askell, Sandhini Agarwal, Ariel Herbert-Voss, Gretchen Krueger, Tom Henighan, Rewon Child, Aditya Ramesh, Daniel~M. Ziegler, Jeffrey Wu, Clemens Winter, Christopher Hesse, Mark Chen, Eric Sigler, Mateusz Litwin, Scott Gray, Benjamin Chess, Jack Clark, Christopher Berner, Sam McCandlish, Alec Radford, Ilya Sutskever, and Dario Amodei.
\newblock Language models are few-shot learners, 2020.

\bibitem{JMLR:v24:22-1144}
Aakanksha Chowdhery, Sharan Narang, Jacob Devlin, Maarten Bosma, Gaurav Mishra, Adam Roberts, Paul Barham, Hyung~Won Chung, Charles Sutton, Sebastian Gehrmann, Parker Schuh, Kensen Shi, Sasha Tsvyashchenko, Joshua Maynez, Abhishek Rao, Parker Barnes, Yi~Tay, Noam Shazeer, Vinodkumar Prabhakaran, Emily Reif, Nan Du, Ben Hutchinson, Reiner Pope, James Bradbury, Jacob Austin, Michael Isard, Guy Gur-Ari, Pengcheng Yin, Toju Duke, Anselm Levskaya, Sanjay Ghemawat, Sunipa Dev, Henryk Michalewski, Xavier Garcia, Vedant Misra, Kevin Robinson, Liam Fedus, Denny Zhou, Daphne Ippolito, David Luan, Hyeontaek Lim, Barret Zoph, Alexander Spiridonov, Ryan Sepassi, David Dohan, Shivani Agrawal, Mark Omernick, Andrew~M. Dai, Thanumalayan~Sankaranarayana Pillai, Marie Pellat, Aitor Lewkowycz, Erica Moreira, Rewon Child, Oleksandr Polozov, Katherine Lee, Zongwei Zhou, Xuezhi Wang, Brennan Saeta, Mark Diaz, Orhan Firat, Michele Catasta, Jason Wei, Kathy Meier-Hellstern, Douglas Eck, Jeff Dean, Slav Petrov, and Noah Fiedel.
\newblock Palm: Scaling language modeling with pathways.
\newblock {\em Journal of Machine Learning Research}, 24(240):1--113, 2023.

\bibitem{deepseekai2025deepseekr1incentivizingreasoningcapability}
DeepSeek-AI, Daya Guo, Dejian Yang, Haowei Zhang, Junxiao Song, Ruoyu Zhang, Runxin Xu, Qihao Zhu, Shirong Ma, Peiyi Wang, Xiao Bi, Xiaokang Zhang, Xingkai Yu, Yu~Wu, Z.~F. Wu, Zhibin Gou, Zhihong Shao, Zhuoshu Li, Ziyi Gao, Aixin Liu, Bing Xue, Bingxuan Wang, Bochao Wu, Bei Feng, Chengda Lu, Chenggang Zhao, Chengqi Deng, Chenyu Zhang, Chong Ruan, Damai Dai, Deli Chen, Dongjie Ji, Erhang Li, Fangyun Lin, Fucong Dai, Fuli Luo, Guangbo Hao, Guanting Chen, Guowei Li, H.~Zhang, Han Bao, Hanwei Xu, Haocheng Wang, Honghui Ding, Huajian Xin, Huazuo Gao, Hui Qu, Hui Li, Jianzhong Guo, Jiashi Li, Jiawei Wang, Jingchang Chen, Jingyang Yuan, Junjie Qiu, Junlong Li, J.~L. Cai, Jiaqi Ni, Jian Liang, Jin Chen, Kai Dong, Kai Hu, Kaige Gao, Kang Guan, Kexin Huang, Kuai Yu, Lean Wang, Lecong Zhang, Liang Zhao, Litong Wang, Liyue Zhang, Lei Xu, Leyi Xia, Mingchuan Zhang, Minghua Zhang, Minghui Tang, Meng Li, Miaojun Wang, Mingming Li, Ning Tian, Panpan Huang, Peng Zhang, Qiancheng Wang, Qinyu Chen, Qiushi Du, Ruiqi Ge, Ruisong
  Zhang, Ruizhe Pan, Runji Wang, R.~J. Chen, R.~L. Jin, Ruyi Chen, Shanghao Lu, Shangyan Zhou, Shanhuang Chen, Shengfeng Ye, Shiyu Wang, Shuiping Yu, Shunfeng Zhou, Shuting Pan, S.~S. Li, Shuang Zhou, Shaoqing Wu, Shengfeng Ye, Tao Yun, Tian Pei, Tianyu Sun, T.~Wang, Wangding Zeng, Wanjia Zhao, Wen Liu, Wenfeng Liang, Wenjun Gao, Wenqin Yu, Wentao Zhang, W.~L. Xiao, Wei An, Xiaodong Liu, Xiaohan Wang, Xiaokang Chen, Xiaotao Nie, Xin Cheng, Xin Liu, Xin Xie, Xingchao Liu, Xinyu Yang, Xinyuan Li, Xuecheng Su, Xuheng Lin, X.~Q. Li, Xiangyue Jin, Xiaojin Shen, Xiaosha Chen, Xiaowen Sun, Xiaoxiang Wang, Xinnan Song, Xinyi Zhou, Xianzu Wang, Xinxia Shan, Y.~K. Li, Y.~Q. Wang, Y.~X. Wei, Yang Zhang, Yanhong Xu, Yao Li, Yao Zhao, Yaofeng Sun, Yaohui Wang, Yi~Yu, Yichao Zhang, Yifan Shi, Yiliang Xiong, Ying He, Yishi Piao, Yisong Wang, Yixuan Tan, Yiyang Ma, Yiyuan Liu, Yongqiang Guo, Yuan Ou, Yuduan Wang, Yue Gong, Yuheng Zou, Yujia He, Yunfan Xiong, Yuxiang Luo, Yuxiang You, Yuxuan Liu, Yuyang Zhou, Y.~X. Zhu,
  Yanhong Xu, Yanping Huang, Yaohui Li, Yi~Zheng, Yuchen Zhu, Yunxian Ma, Ying Tang, Yukun Zha, Yuting Yan, Z.~Z. Ren, Zehui Ren, Zhangli Sha, Zhe Fu, Zhean Xu, Zhenda Xie, Zhengyan Zhang, Zhewen Hao, Zhicheng Ma, Zhigang Yan, Zhiyu Wu, Zihui Gu, Zijia Zhu, Zijun Liu, Zilin Li, Ziwei Xie, Ziyang Song, Zizheng Pan, Zhen Huang, Zhipeng Xu, Zhongyu Zhang, and Zhen Zhang.
\newblock Deepseek-r1: Incentivizing reasoning capability in llms via reinforcement learning, 2025.

\bibitem{edge2025localglobalgraphrag}
Darren Edge, Ha~Trinh, Newman Cheng, Joshua Bradley, Alex Chao, Apurva Mody, Steven Truitt, Dasha Metropolitansky, Robert~Osazuwa Ness, and Jonathan Larson.
\newblock From local to global: A graph rag approach to query-focused summarization, 2025.

\bibitem{erdogan2024tinyagentfunctioncallingedge}
Lutfi~Eren Erdogan, Nicholas Lee, Siddharth Jha, Sehoon Kim, Ryan Tabrizi, Suhong Moon, Coleman Hooper, Gopala Anumanchipalli, Kurt Keutzer, and Amir Gholami.
\newblock Tinyagent: Function calling at the edge, 2024.

\bibitem{erdogan2025planandactimprovingplanningagents}
Lutfi~Eren Erdogan, Nicholas Lee, Sehoon Kim, Suhong Moon, Hiroki Furuta, Gopala Anumanchipalli, Kurt Keutzer, and Amir Gholami.
\newblock Plan-and-act: Improving planning of agents for long-horizon tasks, 2025.

\bibitem{gu2025surveyllmasajudge}
Jiawei Gu, Xuhui Jiang, Zhichao Shi, Hexiang Tan, Xuehao Zhai, Chengjin Xu, Wei Li, Yinghan Shen, Shengjie Ma, Honghao Liu, Saizhuo Wang, Kun Zhang, Yuanzhuo Wang, Wen Gao, Lionel Ni, and Jian Guo.
\newblock A survey on llm-as-a-judge, 2025.

\bibitem{gutiérrez2025ragmemorynonparametriccontinual}
Bernal~Jiménez Gutiérrez, Yiheng Shu, Weijian Qi, Sizhe Zhou, and Yu~Su.
\newblock From rag to memory: Non-parametric continual learning for large language models, 2025.

\bibitem{kim2024llmcompilerparallelfunction}
Sehoon Kim, Suhong Moon, Ryan Tabrizi, Nicholas Lee, Michael~W. Mahoney, Kurt Keutzer, and Amir Gholami.
\newblock An llm compiler for parallel function calling, 2024.

\bibitem{liu2025toolplannertaskplanningclusters}
Yanming Liu, Xinyue Peng, Jiannan Cao, Shi Bo, Yuwei Zhang, Xuhong Zhang, Sheng Cheng, Xun Wang, Jianwei Yin, and Tianyu Du.
\newblock Tool-planner: Task planning with clusters across multiple tools, 2025.

\bibitem{ma2024nonmyopicgenerationlanguagemodels}
Chang Ma, Haiteng Zhao, Junlei Zhang, Junxian He, and Lingpeng Kong.
\newblock Non-myopic generation of language models for reasoning and planning, 2024.

\bibitem{openai2024gpt4ocard}
OpenAI, :, Aaron Hurst, Adam Lerer, Adam~P. Goucher, Adam Perelman, Aditya Ramesh, Aidan Clark, AJ~Ostrow, Akila Welihinda, Alan Hayes, Alec Radford, Aleksander Mądry, Alex Baker-Whitcomb, Alex Beutel, Alex Borzunov, Alex Carney, Alex Chow, Alex Kirillov, Alex Nichol, Alex Paino, Alex Renzin, Alex~Tachard Passos, Alexander Kirillov, Alexi Christakis, Alexis Conneau, Ali Kamali, Allan Jabri, Allison Moyer, Allison Tam, Amadou Crookes, Amin Tootoochian, Amin Tootoonchian, Ananya Kumar, Andrea Vallone, Andrej Karpathy, Andrew Braunstein, Andrew Cann, Andrew Codispoti, Andrew Galu, Andrew Kondrich, Andrew Tulloch, Andrey Mishchenko, Angela Baek, Angela Jiang, Antoine Pelisse, Antonia Woodford, Anuj Gosalia, Arka Dhar, Ashley Pantuliano, Avi Nayak, Avital Oliver, Barret Zoph, Behrooz Ghorbani, Ben Leimberger, Ben Rossen, Ben Sokolowsky, Ben Wang, Benjamin Zweig, Beth Hoover, Blake Samic, Bob McGrew, Bobby Spero, Bogo Giertler, Bowen Cheng, Brad Lightcap, Brandon Walkin, Brendan Quinn, Brian Guarraci, Brian Hsu,
  Bright Kellogg, Brydon Eastman, Camillo Lugaresi, Carroll Wainwright, Cary Bassin, Cary Hudson, Casey Chu, Chad Nelson, Chak Li, Chan~Jun Shern, Channing Conger, Charlotte Barette, Chelsea Voss, Chen Ding, Cheng Lu, Chong Zhang, Chris Beaumont, Chris Hallacy, Chris Koch, Christian Gibson, Christina Kim, Christine Choi, Christine McLeavey, Christopher Hesse, Claudia Fischer, Clemens Winter, Coley Czarnecki, Colin Jarvis, Colin Wei, Constantin Koumouzelis, Dane Sherburn, Daniel Kappler, Daniel Levin, Daniel Levy, David Carr, David Farhi, David Mely, David Robinson, David Sasaki, Denny Jin, Dev Valladares, Dimitris Tsipras, Doug Li, Duc~Phong Nguyen, Duncan Findlay, Edede Oiwoh, Edmund Wong, Ehsan Asdar, Elizabeth Proehl, Elizabeth Yang, Eric Antonow, Eric Kramer, Eric Peterson, Eric Sigler, Eric Wallace, Eugene Brevdo, Evan Mays, Farzad Khorasani, Felipe~Petroski Such, Filippo Raso, Francis Zhang, Fred von Lohmann, Freddie Sulit, Gabriel Goh, Gene Oden, Geoff Salmon, Giulio Starace, Greg Brockman, Hadi
  Salman, Haiming Bao, Haitang Hu, Hannah Wong, Haoyu Wang, Heather Schmidt, Heather Whitney, Heewoo Jun, Hendrik Kirchner, Henrique~Ponde de~Oliveira~Pinto, Hongyu Ren, Huiwen Chang, Hyung~Won Chung, Ian Kivlichan, Ian O'Connell, Ian O'Connell, Ian Osband, Ian Silber, Ian Sohl, Ibrahim Okuyucu, Ikai Lan, Ilya Kostrikov, Ilya Sutskever, Ingmar Kanitscheider, Ishaan Gulrajani, Jacob Coxon, Jacob Menick, Jakub Pachocki, James Aung, James Betker, James Crooks, James Lennon, Jamie Kiros, Jan Leike, Jane Park, Jason Kwon, Jason Phang, Jason Teplitz, Jason Wei, Jason Wolfe, Jay Chen, Jeff Harris, Jenia Varavva, Jessica~Gan Lee, Jessica Shieh, Ji~Lin, Jiahui Yu, Jiayi Weng, Jie Tang, Jieqi Yu, Joanne Jang, Joaquin~Quinonero Candela, Joe Beutler, Joe Landers, Joel Parish, Johannes Heidecke, John Schulman, Jonathan Lachman, Jonathan McKay, Jonathan Uesato, Jonathan Ward, Jong~Wook Kim, Joost Huizinga, Jordan Sitkin, Jos Kraaijeveld, Josh Gross, Josh Kaplan, Josh Snyder, Joshua Achiam, Joy Jiao, Joyce Lee, Juntang
  Zhuang, Justyn Harriman, Kai Fricke, Kai Hayashi, Karan Singhal, Katy Shi, Kavin Karthik, Kayla Wood, Kendra Rimbach, Kenny Hsu, Kenny Nguyen, Keren Gu-Lemberg, Kevin Button, Kevin Liu, Kiel Howe, Krithika Muthukumar, Kyle Luther, Lama Ahmad, Larry Kai, Lauren Itow, Lauren Workman, Leher Pathak, Leo Chen, Li~Jing, Lia Guy, Liam Fedus, Liang Zhou, Lien Mamitsuka, Lilian Weng, Lindsay McCallum, Lindsey Held, Long Ouyang, Louis Feuvrier, Lu~Zhang, Lukas Kondraciuk, Lukasz Kaiser, Luke Hewitt, Luke Metz, Lyric Doshi, Mada Aflak, Maddie Simens, Madelaine Boyd, Madeleine Thompson, Marat Dukhan, Mark Chen, Mark Gray, Mark Hudnall, Marvin Zhang, Marwan Aljubeh, Mateusz Litwin, Matthew Zeng, Max Johnson, Maya Shetty, Mayank Gupta, Meghan Shah, Mehmet Yatbaz, Meng~Jia Yang, Mengchao Zhong, Mia Glaese, Mianna Chen, Michael Janner, Michael Lampe, Michael Petrov, Michael Wu, Michele Wang, Michelle Fradin, Michelle Pokrass, Miguel Castro, Miguel Oom~Temudo de~Castro, Mikhail Pavlov, Miles Brundage, Miles Wang, Minal
  Khan, Mira Murati, Mo~Bavarian, Molly Lin, Murat Yesildal, Nacho Soto, Natalia Gimelshein, Natalie Cone, Natalie Staudacher, Natalie Summers, Natan LaFontaine, Neil Chowdhury, Nick Ryder, Nick Stathas, Nick Turley, Nik Tezak, Niko Felix, Nithanth Kudige, Nitish Keskar, Noah Deutsch, Noel Bundick, Nora Puckett, Ofir Nachum, Ola Okelola, Oleg Boiko, Oleg Murk, Oliver Jaffe, Olivia Watkins, Olivier Godement, Owen Campbell-Moore, Patrick Chao, Paul McMillan, Pavel Belov, Peng Su, Peter Bak, Peter Bakkum, Peter Deng, Peter Dolan, Peter Hoeschele, Peter Welinder, Phil Tillet, Philip Pronin, Philippe Tillet, Prafulla Dhariwal, Qiming Yuan, Rachel Dias, Rachel Lim, Rahul Arora, Rajan Troll, Randall Lin, Rapha~Gontijo Lopes, Raul Puri, Reah Miyara, Reimar Leike, Renaud Gaubert, Reza Zamani, Ricky Wang, Rob Donnelly, Rob Honsby, Rocky Smith, Rohan Sahai, Rohit Ramchandani, Romain Huet, Rory Carmichael, Rowan Zellers, Roy Chen, Ruby Chen, Ruslan Nigmatullin, Ryan Cheu, Saachi Jain, Sam Altman, Sam Schoenholz, Sam
  Toizer, Samuel Miserendino, Sandhini Agarwal, Sara Culver, Scott Ethersmith, Scott Gray, Sean Grove, Sean Metzger, Shamez Hermani, Shantanu Jain, Shengjia Zhao, Sherwin Wu, Shino Jomoto, Shirong Wu, Shuaiqi, Xia, Sonia Phene, Spencer Papay, Srinivas Narayanan, Steve Coffey, Steve Lee, Stewart Hall, Suchir Balaji, Tal Broda, Tal Stramer, Tao Xu, Tarun Gogineni, Taya Christianson, Ted Sanders, Tejal Patwardhan, Thomas Cunninghman, Thomas Degry, Thomas Dimson, Thomas Raoux, Thomas Shadwell, Tianhao Zheng, Todd Underwood, Todor Markov, Toki Sherbakov, Tom Rubin, Tom Stasi, Tomer Kaftan, Tristan Heywood, Troy Peterson, Tyce Walters, Tyna Eloundou, Valerie Qi, Veit Moeller, Vinnie Monaco, Vishal Kuo, Vlad Fomenko, Wayne Chang, Weiyi Zheng, Wenda Zhou, Wesam Manassra, Will Sheu, Wojciech Zaremba, Yash Patil, Yilei Qian, Yongjik Kim, Youlong Cheng, Yu~Zhang, Yuchen He, Yuchen Zhang, Yujia Jin, Yunxing Dai, and Yury Malkov.
\newblock Gpt-4o system card, 2024.

\bibitem{qian2025toolrlrewardtoollearning}
Cheng Qian, Emre~Can Acikgoz, Qi~He, Hongru Wang, Xiusi Chen, Dilek Hakkani-Tür, Gokhan Tur, and Heng Ji.
\newblock Toolrl: Reward is all tool learning needs, 2025.

\bibitem{qin2023toolllmfacilitatinglargelanguage}
Yujia Qin, Shihao Liang, Yining Ye, Kunlun Zhu, Lan Yan, Yaxi Lu, Yankai Lin, Xin Cong, Xiangru Tang, Bill Qian, Sihan Zhao, Lauren Hong, Runchu Tian, Ruobing Xie, Jie Zhou, Mark Gerstein, Dahai Li, Zhiyuan Liu, and Maosong Sun.
\newblock Toolllm: Facilitating large language models to master 16000+ real-world apis, 2023.

\bibitem{sturua2024jinaembeddingsv3multilingualembeddingstask}
Saba Sturua, Isabelle Mohr, Mohammad~Kalim Akram, Michael Günther, Bo~Wang, Markus Krimmel, Feng Wang, Georgios Mastrapas, Andreas Koukounas, Andreas Koukounas, Nan Wang, and Han Xiao.
\newblock jina-embeddings-v3: Multilingual embeddings with task lora, 2024.

\bibitem{kimiteam2025kimik2openagentic}
Kimi Team, Yifan Bai, Yiping Bao, Guanduo Chen, Jiahao Chen, Ningxin Chen, Ruijue Chen, Yanru Chen, Yuankun Chen, Yutian Chen, Zhuofu Chen, Jialei Cui, Hao Ding, Mengnan Dong, Angang Du, Chenzhuang Du, Dikang Du, Yulun Du, Yu~Fan, Yichen Feng, Kelin Fu, Bofei Gao, Hongcheng Gao, Peizhong Gao, Tong Gao, Xinran Gu, Longyu Guan, Haiqing Guo, Jianhang Guo, Hao Hu, Xiaoru Hao, Tianhong He, Weiran He, Wenyang He, Chao Hong, Yangyang Hu, Zhenxing Hu, Weixiao Huang, Zhiqi Huang, Zihao Huang, Tao Jiang, Zhejun Jiang, Xinyi Jin, Yongsheng Kang, Guokun Lai, Cheng Li, Fang Li, Haoyang Li, Ming Li, Wentao Li, Yanhao Li, Yiwei Li, Zhaowei Li, Zheming Li, Hongzhan Lin, Xiaohan Lin, Zongyu Lin, Chengyin Liu, Chenyu Liu, Hongzhang Liu, Jingyuan Liu, Junqi Liu, Liang Liu, Shaowei Liu, T.~Y. Liu, Tianwei Liu, Weizhou Liu, Yangyang Liu, Yibo Liu, Yiping Liu, Yue Liu, Zhengying Liu, Enzhe Lu, Lijun Lu, Shengling Ma, Xinyu Ma, Yingwei Ma, Shaoguang Mao, Jie Mei, Xin Men, Yibo Miao, Siyuan Pan, Yebo Peng, Ruoyu Qin, Bowen Qu, Zeyu
  Shang, Lidong Shi, Shengyuan Shi, Feifan Song, Jianlin Su, Zhengyuan Su, Xinjie Sun, Flood Sung, Heyi Tang, Jiawen Tao, Qifeng Teng, Chensi Wang, Dinglu Wang, Feng Wang, Haiming Wang, Jianzhou Wang, Jiaxing Wang, Jinhong Wang, Shengjie Wang, Shuyi Wang, Yao Wang, Yejie Wang, Yiqin Wang, Yuxin Wang, Yuzhi Wang, Zhaoji Wang, Zhengtao Wang, Zhexu Wang, Chu Wei, Qianqian Wei, Wenhao Wu, Xingzhe Wu, Yuxin Wu, Chenjun Xiao, Xiaotong Xie, Weimin Xiong, Boyu Xu, Jing Xu, Jinjing Xu, L.~H. Xu, Lin Xu, Suting Xu, Weixin Xu, Xinran Xu, Yangchuan Xu, Ziyao Xu, Junjie Yan, Yuzi Yan, Xiaofei Yang, Ying Yang, Zhen Yang, Zhilin Yang, Zonghan Yang, Haotian Yao, Xingcheng Yao, Wenjie Ye, Zhuorui Ye, Bohong Yin, Longhui Yu, Enming Yuan, Hongbang Yuan, Mengjie Yuan, Haobing Zhan, Dehao Zhang, Hao Zhang, Wanlu Zhang, Xiaobin Zhang, Yangkun Zhang, Yizhi Zhang, Yongting Zhang, Yu~Zhang, Yutao Zhang, Yutong Zhang, Zheng Zhang, Haotian Zhao, Yikai Zhao, Huabin Zheng, Shaojie Zheng, Jianren Zhou, Xinyu Zhou, Zaida Zhou, Zhen Zhu,
  Weiyu Zhuang, and Xinxing Zu.
\newblock Kimi k2: Open agentic intelligence, 2025.

\bibitem{touvron2023llama2openfoundation}
Hugo Touvron, Louis Martin, Kevin Stone, Peter Albert, Amjad Almahairi, Yasmine Babaei, Nikolay Bashlykov, Soumya Batra, Prajjwal Bhargava, Shruti Bhosale, Dan Bikel, Lukas Blecher, Cristian~Canton Ferrer, Moya Chen, Guillem Cucurull, David Esiobu, Jude Fernandes, Jeremy Fu, Wenyin Fu, Brian Fuller, Cynthia Gao, Vedanuj Goswami, Naman Goyal, Anthony Hartshorn, Saghar Hosseini, Rui Hou, Hakan Inan, Marcin Kardas, Viktor Kerkez, Madian Khabsa, Isabel Kloumann, Artem Korenev, Punit~Singh Koura, Marie-Anne Lachaux, Thibaut Lavril, Jenya Lee, Diana Liskovich, Yinghai Lu, Yuning Mao, Xavier Martinet, Todor Mihaylov, Pushkar Mishra, Igor Molybog, Yixin Nie, Andrew Poulton, Jeremy Reizenstein, Rashi Rungta, Kalyan Saladi, Alan Schelten, Ruan Silva, Eric~Michael Smith, Ranjan Subramanian, Xiaoqing~Ellen Tan, Binh Tang, Ross Taylor, Adina Williams, Jian~Xiang Kuan, Puxin Xu, Zheng Yan, Iliyan Zarov, Yuchen Zhang, Angela Fan, Melanie Kambadur, Sharan Narang, Aurelien Rodriguez, Robert Stojnic, Sergey Edunov, and Thomas
  Scialom.
\newblock Llama 2: Open foundation and fine-tuned chat models, 2023.

\bibitem{wang2024executablecodeactionselicit}
Xingyao Wang, Yangyi Chen, Lifan Yuan, Yizhe Zhang, Yunzhu Li, Hao Peng, and Heng Ji.
\newblock Executable code actions elicit better llm agents, 2024.

\bibitem{wen2024unlockingreasoningpotentiallarge}
Jiaxin Wen, Jian Guan, Hongning Wang, Wei Wu, and Minlie Huang.
\newblock Unlocking reasoning potential in large langauge models by scaling code-form planning, 2024.

\bibitem{xu2023rewoodecouplingreasoningobservations}
Binfeng Xu, Zhiyuan Peng, Bowen Lei, Subhabrata Mukherjee, Yuchen Liu, and Dongkuan Xu.
\newblock Rewoo: Decoupling reasoning from observations for efficient augmented language models, 2023.

\bibitem{xu2025comprehensivesurveydeepresearch}
Renjun Xu and Jingwen Peng.
\newblock A comprehensive survey of deep research: Systems, methodologies, and applications, 2025.

\bibitem{yang2025reasonfluxhierarchicalllmreasoning}
Ling Yang, Zhaochen Yu, Bin Cui, and Mengdi Wang.
\newblock Reasonflux: Hierarchical llm reasoning via scaling thought templates, 2025.

\bibitem{yao2023treethoughtsdeliberateproblem}
Shunyu Yao, Dian Yu, Jeffrey Zhao, Izhak Shafran, Thomas~L. Griffiths, Yuan Cao, and Karthik Narasimhan.
\newblock Tree of thoughts: Deliberate problem solving with large language models, 2023.

\bibitem{yao2023reactsynergizingreasoningacting}
Shunyu Yao, Jeffrey Zhao, Dian Yu, Nan Du, Izhak Shafran, Karthik Narasimhan, and Yuan Cao.
\newblock React: Synergizing reasoning and acting in language models, 2023.

\bibitem{zeng2023glm130bopenbilingualpretrained}
Aohan Zeng, Xiao Liu, Zhengxiao Du, Zihan Wang, Hanyu Lai, Ming Ding, Zhuoyi Yang, Yifan Xu, Wendi Zheng, Xiao Xia, Weng~Lam Tam, Zixuan Ma, Yufei Xue, Jidong Zhai, Wenguang Chen, Peng Zhang, Yuxiao Dong, and Jie Tang.
\newblock Glm-130b: An open bilingual pre-trained model, 2023.

\bibitem{zhao2025improvinglargelanguagemodel}
Xinran Zhao, Hanie Sedghi, Bernd Bohnet, Dale Schuurmans, and Azade Nova.
\newblock Improving large language model planning with action sequence similarity, 2025.

\end{thebibliography}
\bibliographystyle{plain}

\end{document}